\documentclass{article} 
\usepackage[preprint]{colm2025_conference}

\usepackage{microtype}
\usepackage{hyperref}
\usepackage{url}
\usepackage{booktabs}

\usepackage{lineno}
\usepackage{graphicx}
\usepackage{algorithm}
\usepackage{algpseudocode}
\usepackage{listings}

\usepackage{tcolorbox}
\tcbuselibrary{listingsutf8}
\usepackage{listings}
\usepackage{xcolor}

\definecolor{reply}{HTML}{a020f0}

\definecolor{darkblue}{rgb}{0, 0, 0.5}
\hypersetup{colorlinks=true, citecolor=darkblue, linkcolor=darkblue, urlcolor=darkblue}

\title{PAFFA: Premeditated Actions For Fast Agents}

\author{Shambhavi Krishna \thanks{Work done as part of an internship at Amazon.} \\
University of Massachusetts Amherst \\
\texttt{shambhavikri@umass.edu} \\
\And
Zheng Chen\\
Amazon Alexa\\
\texttt{tzchen86@gmail.com} \\
\And
Yuan Ling \\
Amazon Benchmarking \\
\And
Xiaojiang Huang \\
Amazon Alexa \\
\And
Yingjie Li \\
Amazon Alexa \\
\And
Fan Yang \\
Amazon Alexa \\
\And
Xiang Li \\
Amazon Alexa \\
}

\begin{document}

\ifcolmsubmission
\linenumbers
\fi

\maketitle

\begin{abstract}
Modern AI assistants have made significant progress in natural language understanding and tool-use, with emerging efforts to interact with Web interfaces. However, current approaches that heavily rely on repeated LLM-driven HTML parsing are computationally expensive and error-prone, particularly when handling dynamic web interfaces and multi-step tasks. We introduce PAFFA (Premeditated Actions For Fast Agents), a method that makes LLMs faster and more accurate in completing tasks on the internet using a novel inference-time technique that requires no task-specific training. PAFFA constructs an ``Action Library'', leveraging the parametric knowledge of the base LLM to pre-compute browser interaction patterns that generalize across tasks. By strategically re-using LLM inference \textit{across} tasks --- either via "Dist-Map" for task-agnostic identification of key interactive web elements, or "Unravel" for first-encounter, stateful exploration of novel tasks/sites) --- PAFFA drastically reduces inference time tokens by 87\% while maintaining robust performance (achieving 0.57 vs. 0.50 step accuracy compared to baseline). Further, Unravel's ability to update its action library based on explorations allows generalization and adaptation to unseen websites. In sum, this work exhibits that LLM reasoning sequences can generalize across prompts, offering a way to scale inference-time techniques for internet-scale data with sublinear token count.
\end{abstract}



\section{Introduction}
AI Assistants are increasingly expected to navigate the Internet autonomously to complete user tasks~\citep{he2024webvoyager, deng2023mind2webgeneralistagentweb}. While AI has advanced in natural language understanding and using structured APIs, interacting reliably with diverse, dynamic web interfaces \footnote{i.e., interfaces where content, structure, or element identifiers frequently change} remains a major challenge, limiting the practical use of these agents~\citep{liu2018reinforcement}. The web's semi-structured, visual, and ever-changing nature presents distinct hurdles compared to controlled environments. Effective web interaction requires agents to parse HTML/DOM structure, connect language instructions to visual elements, plan action sequences, and adapt robustly to interface changes - capabilities that challenge current models.

Many current web agents follow an approach requiring repeated computation: they invoke Large Language Models (LLMs) at each step to parse the full HTML and decide the next action~\citep{mazumder2020flin, lù2024weblinxrealworldwebsitenavigation, deng2023mind2webgeneralistagentweb}. While this uses the LLM's strong language understanding, this method has significant drawbacks for web interaction:

\begin{itemize}
    \item \textbf{Efficiency:} Repetitively parsing complex DOMs is computationally expensive.
    
    \item \textbf{Reliability:} Web interfaces change constantly. Agents relying on direct, step-by-step HTML parsing are brittle; small DOM changes can cause action failures and errors~\citep{pan2024autonomous}.
    
    \item \textbf{Scalability:} Solutions often require website-specific implementations or struggle with the diversity of the web. Furthermore, handling multi-page tasks can quickly exceed LLM context limits when processing cumulative HTML, limiting agent versatility and the complexity of tasks they can handle reliably.
    
\end{itemize}

To address these issues, we introduce PAFFA (Premeditated Actions For Fast Agents). PAFFA offers an alternative to repetitive runtime parsing through a structured, efficient method grounded in pre-computation of inference and strategic LLM application. PAFFA's core is an Action Library: a persistent library of reusable, parameterized functions encoding verified interaction patterns for websites. This library effectively caches the results of complex reasoning and planning required for specific web interactions. 

This library is primarily constructed offline, leveraging the inherent parametric knowledge and zero-shot capabilities of base LLMs without requiring task-specific fine-tuning or expert demonstrations. PAFFA offers two distinct strategies for this initial library generation:

\begin{itemize}
    \item \textbf{Dist-Map:} This strategy first performs task-agnostic element distillation using LLM semantic understanding to extract key interactive elements into a simplified, robust representation. Subsequently, task-specific scripts (forming the basis of APIs) are generated using these distilled elements. This approach aims for efficiency and robustness to UI variation by abstracting common structures.
    \item \textbf{Unravel (for Construction):} Alternatively, Unravel can directly generate task-specific interaction scripts (which are then parameterized into actions) by processing tasks incrementally, using the full HTML context one page/state at a time. This method handles complex interactions directly without prior distillation.
\end{itemize}

Regardless of the initial construction strategy, at runtime, PAFFA employs a novel inference-time technique: the LLM performs lightweight high-level intent recognition (matching user requests to APIs) and parameter extraction, followed by direct execution of the corresponding pre-computed API from the library. This division of labor greatly reduces costly real-time HTML parsing.

Furthermore, the Unravel methodology possesses a unique dual capability. Beyond its potential use in initial library construction, Unravel serves as PAFFA's primary mechanism for dynamic adaptation and handling novelty. When encountering a task, website, or significant structural change not covered by the existing Action library, PAFFA invokes Unravel to handle the interaction live, performing stateful exploration on the current interface. This needs to be done only once for the first encounter with that specific new scenario.

Crucially, the interaction sequences successfully executed by Unravel during these first encounters can be captured, analyzed, parameterized, and integrated back into the Action Library. This update mechanism allows PAFFA to continuously learn, adapt, and generalize from new experiences, effectively caching solutions for novel or modified interactions and maintaining effectiveness as websites evolve – all without retraining the base LLM.

Through these methods, PAFFA demonstrates significant benefits. Evaluations on Mind2Web show \textbf{superior performance} over baselines (e.g. element accuracy: 0.74 vs 0.56; step accuracy: 0.57 vs 0.50 on Air.+All Shop. set - see Section~\ref{sec:results}) while drastically \textbf{reducing computational cost}, with an 87\% reduction in inference tokens during execution compared to typical LLM-parsing methods.

This work offers:
\begin{itemize}
    \item A novel Action Library paradigm that pre-computes and caches web interaction patterns, minimizing runtime computation and leveraging LLM parametric knowledge offline without training.
    \item Two distinct offline strategies (Dist-Map via distillation, Unravel via incremental exploration) for initial library construction using zero-shot LLMs.
    \item Unravel's dual role in enabling both initial generation and, uniquely, dynamic runtime adaptation to novel scenarios (new tasks/sites/changes) coupled with a mechanism for library evolution and generalization.
    \item Empirical results showing significant efficiency gains and strong accuracy improvements on the Mind2Web web agent benchmark.
\end{itemize}

By replacing expensive runtime parsing with the lightweight execution of pre-computed action plans stored in an adaptive library, PAFFA provides a more scalable, efficient, and robust path for capable web agents. This work contributes to optimizing language model use in complex interactive domains by demonstrating effective strategies for caching, generalizing, and adapting complex reasoning processes.


\section{Related Works}
\label{sec:related_work}

\textbf{Foundation Models and Web Interaction Frameworks: } Early web automation frameworks like Mini-WoB++~\citep{he2024webvoyager} provided controlled environments for basic web tasks, but failed to capture real-world complexity. Mind2Web~\citep{deng2023mind2webgeneralistagentweb} advanced the field by incorporating real-world websites, though challenges remain in handling dynamic content. Pre-trained models have shown particular promise, with bidirectional models like HTML-T5 achieving state-of-the-art results in document parsing~\citep{li2021markuplm}.

\textbf{Multimodal and Vision-Based Approaches: } Recent work has explored multimodal interactions for web navigation. WebVoyager~\citep{he2024webvoyager} leverages multimodal models for understanding both visual and textual elements, while Pix2Act~\citep{shaw2023pixels} demonstrates success in screenshot parsing and behavioral cloning using Monte Carlo Tree Search.

\textbf{Navigation and Planning Systems: } Frameworks like FLIN~\citep{mazumder2020flin} and WebLINX~\citep{lù2024weblinxrealworldwebsitenavigation} have advanced natural language-based navigation, though their step-by-step planning mechanisms create efficiency bottlenecks. Recent work~\citep{gur2023real} has shown promise in task decomposition and multi-step interactions, while MindSearch~\citep{ma2023laser} introduces graph-based planning strategies.

\textbf{Action Abstraction and Evaluation: } Current approaches face challenges in balancing accuracy with computational efficiency, requiring repeated HTML parsing and LLM inference. Mind2Web-Live~\citep{pan2024webcanvas} introduces progress-aware evaluation allowing multiple valid paths to task completion. While frameworks have attempted to create reusable components, most focus on low-level actions. Recent work in self-experience supervision~\citep{gur2023real} and frameworks like TPTU-v2~\citep{kong2023tptu} explore promising directions in tool use and planning, though gaps remain in developing flexible, high-level action APIs.

\section{Methodology}
\label{sec:methodology}

\subsection{Overview: Shifting away solely per-interaction reasoning/action sequences}

This section details the PAFFA framework, outlining its core components and processes designed to enable efficient, robust, and adaptable web interaction without task-specific training. PAFFA centers around an \textbf{Action Library} built using Large Language Models (LLMs) via zero-shot prompting.

We first describe the \textbf{two alternative strategies} for initial library construction: \textbf{Dist-Map}, which relies on element distillation, and \textbf{Unravel}, which performs direct incremental generation (Section~\ref{sec:initial_construction}). We then detail \textbf{Unravel's unique role in runtime adaptation} to novel scenarios and the mechanism for \textbf{evolving the Action API Library} based on these runtime experiences (Section~\ref{subsec:runtime}). Finally, we cover the common processes of \textbf{API synthesis} from generated scripts/traces and the \textbf{runtime execution model} using the library (Section~\ref{subsec:lib}).




\begin{figure}[htbp]
    \centering
    \begin{minipage}{0.48\textwidth}
        \centering
        \includegraphics[width=\textwidth]{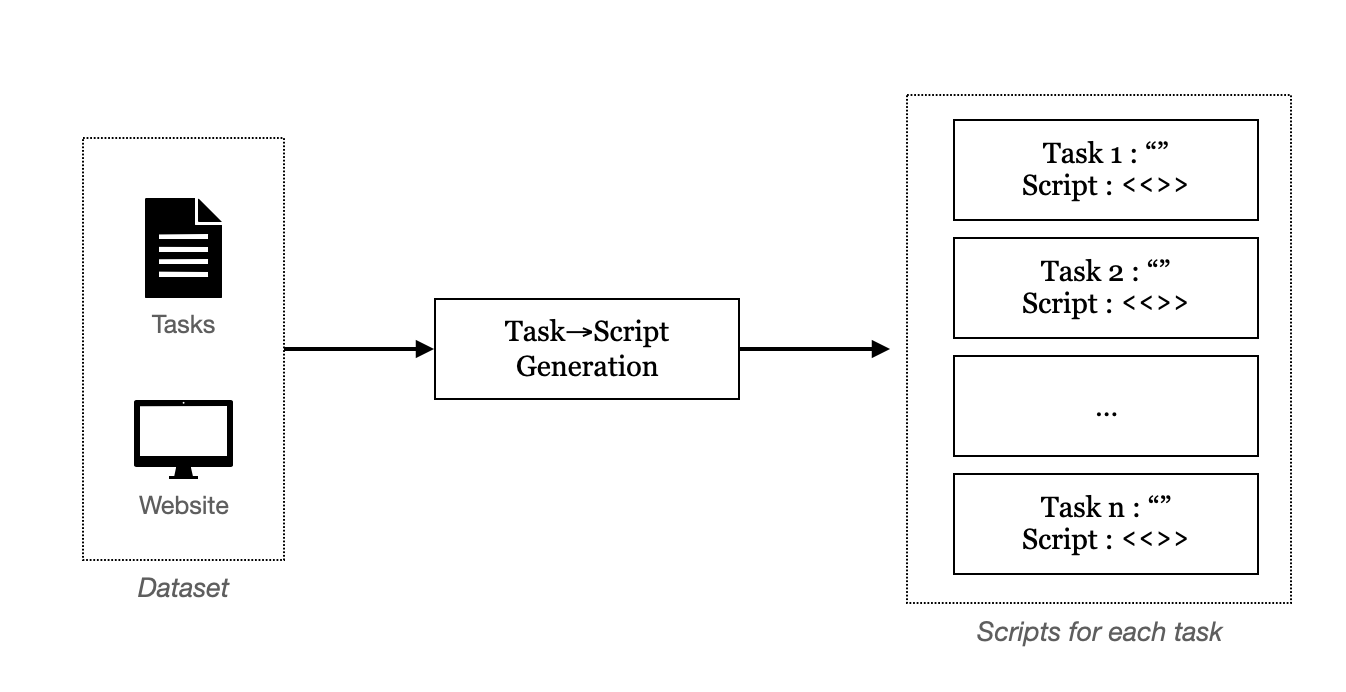}
        \caption{Creating task-specific scripts.}
        \label{fig:step1a}
    \end{minipage}\hfill
    \begin{minipage}{0.48\textwidth}
        \centering
        \includegraphics[width=\textwidth]{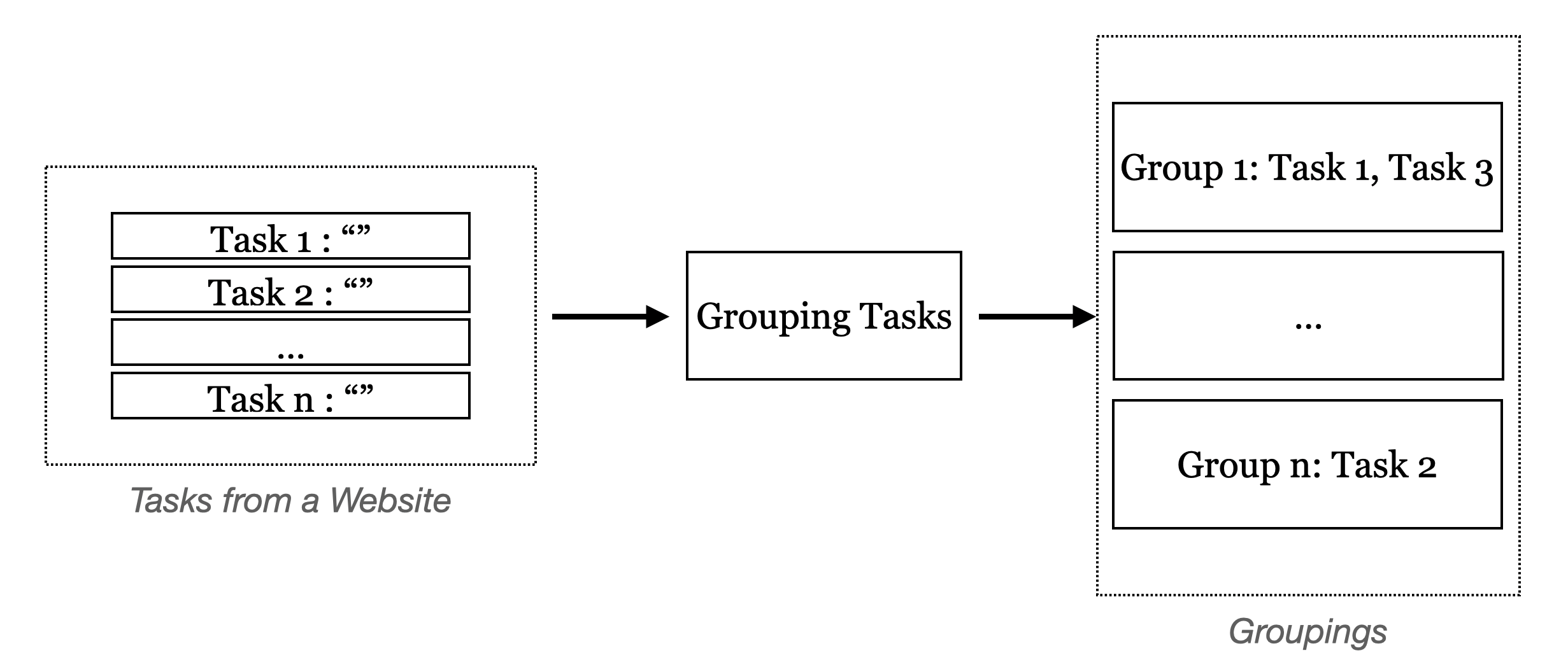}
        \caption{Grouping tasks solvable by one API.}
        \label{fig:step1b}
    \end{minipage}

    \vspace{0.5cm} 

    \begin{minipage}{0.68\textwidth}
        \centering
        \includegraphics[width=\textwidth]{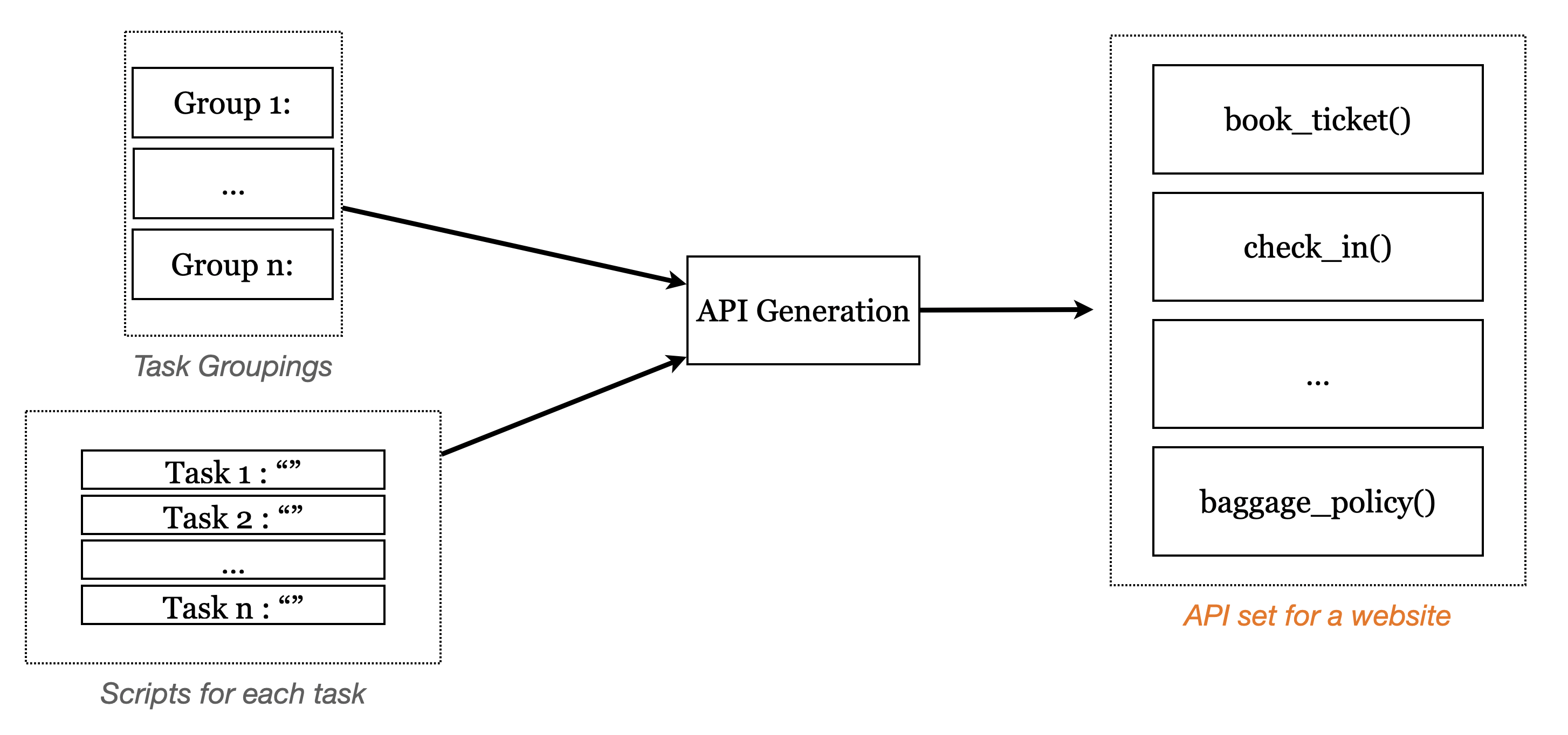}
        \caption{Creating APIs per group.}
        \label{fig:step1c}
    \end{minipage}
\end{figure}

The Action Library is constructed by leveraging the inherent document understanding, reasoning, and code generation capabilities of LLMs \footnote{All LLM operations described in this section utilize Anthropic's Sonnet 3.5~\citep{anthropic2025claude} model.}.

\subsection{Initial Library Construction Strategies}
\label{sec:initial_construction}

\subsubsection{Strategy 1: Dist-Map (Distillation then Mapping)}

The goal of this method is to search for reusable actions that will generalize across interface variations. We hypothesize that leveraging the LLM's inherent understanding of webpages to ``distill'' the semantic meaning of page elements will make the agent invariant to superficial UI changes.  Then, we leverage the LLM's coding ability to write scripts that operate on the distilled semantic representations of webpage elements.  More specifically, we break the algorithm into 3 parts:

\begin{itemize}
    \item Phase 1: Task-Agnostic Element Distillation:  Use the LLM for targeted semantic analysis of HTML to identify and extract key interactive elements (buttons, fields, links) and map them to their attributes, creating a structured, distilled representation (e.g., JSON). This aims to capture the functional essence independent of precise layout, building upon work from ~\citep{gur2023real,zheng2023synapse}.

    \item Phase 2: Verification: Ask the LLM to review the distilled elements to ensure their correctness.
    
    \item Phase 3: Contextual Script Generation: Map a specific task description to a sequence of interactions using only the relevant verified, distilled elements. The LLM generates an executable Selenium script~\citep{selenium-webdriver} based on this constrained context. 
\end{itemize}

Because the action library constructed by Dist-Map is static, this algorithm is well-suited for stable, frequently used interaction patterns where element abstraction provides robustness.

\subsubsection{Strategy 2: Unravel (Direct Incremental Exploration)}
\label{subsub:unravel}
This methodology directly generates scripts by simulating the interaction incrementally, leveraging the full context of the current page state at each step.

For a given task description, Unravel decomposes it into sequential subtasks corresponding to interactions within single page views. In a loop, the LLM processes the \textit{full HTML} of the current page, along with the \textbf{overall task goal and interaction history}. Unlike methods that greedily select only the single next best action, this richer context allows the LLM to perform more sophisticated localized planning -- potentially identifying multiple necessary actions on the current page or generating code that implicitly anticipates subsequent steps within the current view -- before generating the code for the immediate interaction(s).

By using the full page context per step, we observe that this method handles more complex interactions than Dist-Map with more robust error handling (see Section~\ref{subsubsec:novelty}) in the initial generated action scripts. It avoids the potential information loss of distillation but processes more raw HTML during script generation compared to Dist-Map's final phase.
    
\subsection{Runtime Adaptation and Library Evolution via Unravel}
\label{subsec:runtime}
A unique aspect of PAFFA is its ability to adapt to novelty and evolve its library over time, primarily driven by the Unravel methodology operating at runtime.
\subsubsection{Handling Novelty with Unravel}
\label{subsubsec:novelty}
When a user request corresponds to a task, website, or significantly changed interface not covered by the existing Action Library, PAFFA invokes Unravel dynamically:
\begin{itemize}
    \item Trigger: The initial API retrieval fails, or a previously reliable API encounters repeated execution errors indicative of major site changes.
    \item Execution: Unravel engages in incremental, stateful exploration, using the process described in Section~\ref{subsub:unravel} (chunked execution based on full current-page HTML, maintaining task coherence via history) to attempt the task live on the interface. This exploration needs to be performed only once for the first successful completion of that specific novel scenario.
    \item Robustness during Exploration: As Unravel processes the full page context at each step, it can generate interaction code with sophisticated error handling (e.g., multi-step try-except blocks targeting different selectors), increasing the chance of success even on unfamiliar or dynamic interfaces.
\end{itemize}

The successful execution traces generated by Unravel during runtime adaptation are valuable new interaction patterns. PAFFA incorporates a mechanism to integrate this knowledge back into the library:
\begin{itemize}
    \item Capture: Record the scripts used by Unravel for a novel task completion.
    \item Synthesize: Process this trace using API Synthesis (Section~\ref{subsec:lib}) to create a new, parameterized API.
    \item Integrate: Add the new API to the Action Library.
\end{itemize}

This feedback loop allows PAFFA to continuously learn and generalize. Solutions discovered for novel scenarios are effectively cached as new APIs, making future execution of the same or similar tasks highly efficient. This adaptation occurs without retraining the base LLM, relying instead on capturing and structuring the results of its runtime problem-solving.

\subsubsection{Library Evolution}
\label{subsec:lib}
Once initial scripts are generated (by either Dist-Map or Unravel) or new traces are captured (from runtime Unravel \footnote{This would be after a full task has been successfully completed by Unravel}), the following steps finalize the API library and enable runtime use:
\begin{enumerate}
    \item Task Clustering: All scripts/traces for a website are used to prompt the model to reason and identify groups of tasks sharing common interaction sequences or sub-goals based on semantic similarity (e.g., various login flows, different search types). This LLM-driven step ensures related functionalities are grouped before parameterization.
    \item API Synthesis and Parameterization: For each task cluster, the LLM performs program/API synthesis. It analyzes the associated scripts/traces, identifies variations (e.g., different search terms, credentials, dates), and generates a single parameterized Python function (an API) capable of executing all tasks within the cluster. For increased robustness and guarantee of task completion, we employ a 2-step reasoning-based self-corrective prompt that is directed to list any possible shortcomings of the script, and then to address them fully. 
\end{enumerate}

\subsubsection{Runtime Usage}
During deployment (Figure \ref{fig:step4} in Appendix~\ref{sec:applib}), when a user request arrives:
\begin{itemize}
    \item API Retrieval/Intent Recognition: The LLM selects the most appropriate API from the library based on the user's natural language request.
    \item Parameter Extraction/Slot Filling: The LLM extracts necessary arguments from the request to call the selected API.
    \item Execution: The chosen, parameterized API is executed directly using browser automation (e.g., Selenium).
\end{itemize}

This runtime workflow replaces expensive, iterative full-page HTML parsing with a lightweight API call, leveraging the pre-computed knowledge encoded in the action library. 

\section{Results}
\label{sec:results}
We evaluate PAFFA's effectiveness through multiple lenses: performance on standard web agent benchmarks, computational efficiency during task execution, and quality of the constructed action library.

\subsection{Mind2Web Benchmark Performance}
\label{subsec:bench}
To compare PAFFA against established benchmarks, we evaluate on the Mind2Web dataset~\citep{deng2023mind2webgeneralistagentweb}, which is the predominant web agent benchmark (to the authors' knowledge). We report performance using two standard metrics:

\begin{enumerate}
    \item Element Accuracy: The percentage of correctly identified interactive elements.
    \item Step Accuracy: The percentage of correct element-action pairs executed.
\end{enumerate}

We follow the Mind2Web evaluation protocol, considering various cross-task and cross-website splits across the Airlines and Shopping domains (details in Appendix~\ref{subsec:dataset_app}). Crucially, web task annotations can be ambiguous, and multiple valid interaction paths often exist. Therefore, in addition to exact match accuracy (`Exact'), we perform human re-evaluation to account for functionally correct but non-annotated paths (`Inexact'), providing a more realistic measure of task success.

Tables~\ref{tab:ele_acc} and \ref{tab:step_acc} present the Element and Step Accuracy results, comparing both PAFFA algorithms (Dist-Map and Unravel) on a Sonnet 3.5 base against MindAct baselines (fine-tuned DeBERTa+FLAN-T5, and Sonnet 3.5 with 3-shot prompting\footnote{They follow the same three-shot strategy for GPT 3.5 and GPT4 in their paper}). They show that PAFFA's methodologies generally achieve superior or competitive performance compared to MindAct baselines across various splits.

\begin{table*}[t]
\footnotesize
\centering
\begin{tabular}{|l|c|c|c|c|c|cc|cc|}
\hline
& \multicolumn{5}{c|}{\textbf{MindAct}} & \multicolumn{2}{c|}{\textbf{Dist-Map}} & \multicolumn{2}{c|}{\textbf{Unravel}} \\
\cline{2-10}
\textbf{Dataset Split} & \textbf{DeB} & \textbf{T5 B} & \textbf{T5 L} & \textbf{T5 XL} & \textbf{S3.5} & \textbf{Ex} & \textbf{Inex} & \textbf{Ex} & \textbf{Inex} \\
\hline
Airlines & 0.306 & 0.477 & 0.562 & 0.626 & 0.492 & 0.618 & 0.67 & 0.75 & \textbf{0.76} \\
Air.+Shop.-Task & 0.285 & 0.428 & 0.508 & 0.566 & 0.418 & 0.699 & 0.78 & 0.67 & \textbf{0.701} \\
Shop.-Cross Task & 0.202 & 0.273 & 0.333 & 0.357 & 0.202 & 0.779 & \textbf{0.89} & 0.59 & 0.642 \\
Shop.-Task+CW & 0.283 & 0.372 & 0.45 & 0.472 & 0.316 & 0.74 & \textbf{0.86} & 0.61 & 0.67 \\
Shop.-Cross-Web & 0.354 & 0.510 & 0.552 & 0.552 & 0.395 & 0.7 & \textbf{0.83} & 0.62 & 0.72 \\
Air.+All Shop. & 0.298 & 0.449 & 0.526 & 0.562 & 0.422 & 0.699 & \textbf{0.79} & 0.65 & 0.74 \\
\hline
\end{tabular}
\caption{\label{tab:ele_acc}Element Accuracies: MindAct vs. PAFFA}
\end{table*}

\begin{table*}[t]
\footnotesize
\centering
\begin{tabular}{|l|c|c|c|c|cc|cc|}
\hline
& \multicolumn{4}{c|}{\textbf{MindAct}} & \multicolumn{2}{c|}{\textbf{Dist-Map}} & \multicolumn{2}{c|}{\textbf{Unravel}} \\
\cline{2-9}
\textbf{Dataset Split} & \textbf{T5 B} & \textbf{T5 L} & \textbf{T5 XL} & \textbf{S3.5} & \textbf{Ex} & \textbf{Inex} & \textbf{Ex} & \textbf{Inex} \\
\hline
Airlines & 0.462 & 0.564 & \textbf{0.616} & 0.467 & 0.35 & 0.42 & 0.38 & 0.58 \\
Air.+Shop.-Task & 0.421 & 0.503 & 0.548 & 0.38 & 0.35 & 0.525 & 0.29 & \textbf{0.55} \\
Shop.-Cross Task & 0.276 & 0.331 & 0.338 & 0.173 & 0.35 & \textbf{0.63} & 0.2 & 0.52 \\
Shop.-Task+CW & 0.286 & 0.366 & 0.384 & 0.231 & 0.344 & 0.60 & 0.29 & \textbf{0.56} \\
Shop.-Cross-Web & 0.357 & 0.385 & 0.362 & 0.249 & 0.33 & 0.56 & 0.36 & \textbf{0.58} \\
Air.+All Shop. & 0.409 & 0.482 & 0.500 & 0.357 & 0.34 & \textbf{0.57} & 0.32 & \textbf{0.57} \\
\hline
\end{tabular}
\caption{\label{tab:step_acc}Step Accuracy: MindAct vs. PAFFA}
\end{table*}
First, comparing PAFFA (using zero-shot Sonnet 3.5) against the fine-tuned MindAct baselines (DeBERTa~\citep{he2021debertadecodingenhancedbertdisentangled} + FLAN-T5~\citep{chung2022scalinginstructionfinetunedlanguagemodels} models up to 3B parameters), PAFFA demonstrates strong performance despite using no task-specific training whatsoever. While Sonnet 3.5 is a larger, more capable base model, PAFFA's effectiveness coupled with its massive reduction in runtime computation (Section~\ref{subsec:cost}) highlights the power of its pre-computation and strategic LLM usage.

Second, to isolate the contribution of the PAFFA algorithm itself from the underlying model's power, we compare PAFFA (zero-shot Sonnet 3.5) against MindAct using the same Sonnet 3.5 model with 3-shot prompting (S3.5 column), following the few-shot strategy used in the original MindAct paper. Even under this controlled comparison using the identical base model, PAFFA often outperforms the MindAct few-shot approach, particularly evident in the more forgiving `Inexact' metrics. This strongly suggests that PAFFA's architectural design -- the Action library built via Dist-Map and Unravel -- provides significant advantages beyond just leveraging a powerful LLM.

Third, the Cross-Website generalization challenge starkly highlights PAFFA's advantage. As seen in the Shop.-Cross-Web split, the performance of MindAct (both fine-tuned and few-shot Sonnet) degrades significantly when faced with websites unseen during training or prompting setup. PAFFA's Unravel method, however, maintains performance comparable to its other splits. This is a direct consequence of PAFFA's design: because PAFFA requires no training data and Unravel adapts dynamically using the base LLM's generalization capabilities, it does not suffer from the same generalization gap when encountering new website structures.

Therefore, PAFFA not only demonstrates superior performance in several scenarios but achieves this without needing fine-tuning data or expert demonstrations. Its training-free nature makes it inherently more robust to domain shifts, like encountering new websites, a critical advantage for real-world deployment.


\subsection{Cost Comparison}
\label{subsec:cost}

A core motivation for PAFFA is reducing the computational cost associated with repeated LLM calls for HTML parsing. We compare the typical inference cost per task at deployment time. MindAct requires stepwise LLM calls involving ranked candidate elements\footnote{This is with k=50 as is preferred in the Mind2Web paper.} and action selection. PAFFA, leveraging its Action Library, requires only a single, simpler LLM call to map the user request to the appropriate pre-computed API and extract parameters.


\begin{table}[]
\resizebox{\columnwidth}{!}{%
\begin{tabular}{|ccl|ccl|}
\hline
\multicolumn{3}{|c|}{\textbf{MindAct}}                                                                        & \multicolumn{3}{c|}{\textbf{PAFFA}}                                                                         \\ \hline
\multicolumn{1}{|c|}{\textbf{\# Tokens}} & \multicolumn{1}{c|}{\textbf{Calls}} & \textbf{Total Tokens}                 & \multicolumn{1}{c|}{\textbf{\# Tokens}} & \multicolumn{1}{c|}{\textbf{Calls}} & \textbf{Total Tokens}       \\ \hline
\multicolumn{1}{|c|}{1,565}              & \multicolumn{1}{c|}{126}            & \multicolumn{1}{c|}{197,190} & \multicolumn{1}{c|}{25,000}             & \multicolumn{1}{c|}{1}              & \multicolumn{1}{c|}{\textbf{25,000}} \\ \hline
\end{tabular}%
}
\caption{After Setup Usage Per Task/Request}
\label{tab:deployment_cost}
\end{table}

PAFFA achieves an 87\% reduction in estimated inference tokens per task compared to the MindAct workflow. This substantial efficiency gain stems directly from replacing expensive, iterative runtime parsing and planning with lightweight calls to pre-computed, verified action primitives stored in the Action Library. This calculation is conservative, as it doesn't include the token cost associated with MindAct's candidate element ranking performed by DeBERTa.

\subsection{Qualitative Evaluation of Generated Actions}
Furthermore, to evaluate the generalization and coverage of the generated Action Library beyond the initial set of tasks, we employed LLM-based synthetic task generation and evaluated the system's ability to correctly ground these diverse requests to the appropriate API calls (see Appendix~\ref{app:api_example} for examples).

Beyond task success rates and computational cost, we analyze the intrinsic quality of the initial scripts generated by PAFFA's core methodologies, Dist-Map and Unravel, before potential API synthesis. We employ LLM-based evaluation using Sonnet 3.5\footnote{We employ Sonnet 3.5 for qualitative assessment, providing consistent evaluations at scale, though we acknowledge this is an emerging evaluation technique.} on the Airlines dataset subset. Scripts were assessed along three dimensions: Script Task Alignment (functional goal completion vs. task requirements), Action Representation Fidelity (faithfulness of code to necessary actions), and Script Efficiency (directness of the interaction path). Detailed definitions and LLM prompts for these metrics are provided in Appendix~\ref{sec:qualapp}\footnote{Since PAFFA does not use any expert trajectories/action sequences, expert actions mentioned in Appendix~\ref{sec:qualapp} prompts serve only as an evaluation reference standard.}.


\textbf{Findings}: Table \ref{tab:qual_eval} summarizes the average scores, comparing scripts generated via the Dist-Map strategy versus the Unravel strategy. Unravel consistently outperforms Dist-Map across all metrics.

\begin{table}[ht]
\centering
\caption{LLM-based Qualitative Evaluation Scores (Avg. on Airlines Dataset). Higher is better.}
\label{tab:qual_eval}
\begin{tabular}{lccc}
\toprule
\textbf{Metric} & \textbf{Dist-Map }& \textbf{Unravel} & \textbf{Rel. Improv.} \\ 
\midrule
Script Task Alignment (1-5) & 3.2 & 3.8 & +18.8\% \\ 
Action Repr. Fidelity (1-5) & 2.1 & 2.77 & +31.9\% \\ 
Script Efficiency (1-5) & 2.1 & 3.0 & +42.9\% \\ 
\bottomrule
\end{tabular}
\end{table}


The qualitative evaluation consistently favors scripts generated by Unravel over Dist-Map across all measured dimensions. The significant relative improvements, particularly in efficiency (+42.9\%) and fidelity (+31.9\%), strongly suggest that Unravel's direct, incremental exploration using full page context produces higher-quality initial scripts. Based on these evaluated metrics, Unravel appears superior for generating effective, well-aligned, and relatively efficient scripts in this domain, reinforcing its importance not only for runtime adaptation (Section~\ref{subsec:runtime}) but also as a strong candidate for the initial library construction phase.



\section{Discussion and Conclusion}
\label{sec:discussion}

This work introduced PAFFA, an algorithm designed to make LLM-driven web agents more efficient, robust and adaptable. By leveraging an Action Library to pre-compute and cache common interaction patterns offline using zero-shot LLMs, PAFFA significantly reduces the computational burden at runtime. Compared to methods requiring step-by-step HTML parsing and planning, PAFFA employs a lightweight inference-time technique involving API retrieval and execution. Our empirical results demonstrate PAFFA's effectiveness: it achieves superior or competitive accuracy on the Mind2Web benchmark while realizing an 87\% reduction in inference tokens during task execution.

A key aspect of PAFFA is its training-free nature. Both initial library construction strategies (Dist-Map and Unravel) rely solely on the parametric knowledge of base LLMs, eliminating the need for costly fine-tuning or expert demonstrations. This inherent characteristic, coupled with the Unravel methodology for runtime adaptation, grants PAFFA strong generalization capabilities, particularly evident in its robust performance on cross-website tasks where traditional trained models falter. Unravel's ability to handle novel scenarios dynamically and feed successful execution traces back into the library allows PAFFA to continuously adapt and evolve without retraining. Our qualitative evaluations further suggest that Unravel, when used for initial script generation, produces higher-fidelity and more efficient code than the distillation-based Dist-Map approach, reinforcing its central role in the framework.

While PAFFA demonstrates significant advantages, we acknowledge limitations. Our current evaluation relies partly on LLM-based qualitative metrics, an emerging technique, and human evaluation for `Inexact' accuracy, which can be resource-intensive. The scope of tested websites and tasks, while based on a standard benchmark, is not exhaustive. Furthermore, while Unravel's design promotes robustness, dedicated empirical evaluation of resilience to various types of website changes is needed. Future work should focus on automating API grouping and grounding, enhancing verification modules for library entries, developing strategies for automated library maintenance (e.g., detecting and updating stale APIs), and integrating PAFFA's capabilities within broader conversational AI assistants.

In conclusion, the Action API Library concept represents a fundamental shift from purely reactive, iterative LLM application towards leveraging pre-computation and strategic caching of complex reasoning processes. PAFFA demonstrates that this approach can yield substantial gains in efficiency and adaptability for autonomous web agents, offering a promising direction for scaling LLM capabilities in complex, dynamic interactive domains.

\bibliography{colm2025_conference}

\begin{thebibliography}{17}
\providecommand{\natexlab}[1]{#1}
\providecommand{\url}[1]{\texttt{#1}}
\expandafter\ifx\csname urlstyle\endcsname\relax
  \providecommand{\doi}[1]{doi: #1}\else
  \providecommand{\doi}{doi: \begingroup \urlstyle{rm}\Url}\fi

\bibitem[sel()]{selenium-webdriver}
Webdriver.
\newblock URL \url{https://www.selenium.dev/documentation/webdriver/}.
\newblock Accessed: 2024-12-02.

\bibitem[Anthropic(2025)]{anthropic2025claude}
Anthropic.
\newblock Claude 3.5 sonnet.
\newblock \url{https://www.anthropic.com}, 2025.
\newblock Large language model developed by Anthropic.

\bibitem[Chung et~al.(2024)Chung, Hou, Longpre, Zoph, Tay, Fedus, Li, Wang, Dehghani, Brahma, Webson, Gu, Dai, Suzgun, Chen, Chowdhery, Castro{-}Ros, Pellat, Robinson, Valter, Narang, Mishra, Yu, Zhao, Huang, Dai, Yu, Petrov, Chi, Dean, Devlin, Roberts, Zhou, Le, and Wei]{chung2022scalinginstructionfinetunedlanguagemodels}
Hyung~Won Chung, Le~Hou, Shayne Longpre, Barret Zoph, Yi~Tay, William Fedus, Yunxuan Li, Xuezhi Wang, Mostafa Dehghani, Siddhartha Brahma, Albert Webson, Shixiang~Shane Gu, Zhuyun Dai, Mirac Suzgun, Xinyun Chen, Aakanksha Chowdhery, Alex Castro{-}Ros, Marie Pellat, Kevin Robinson, Dasha Valter, Sharan Narang, Gaurav Mishra, Adams Yu, Vincent~Y. Zhao, Yanping Huang, Andrew~M. Dai, Hongkun Yu, Slav Petrov, Ed~H. Chi, Jeff Dean, Jacob Devlin, Adam Roberts, Denny Zhou, Quoc~V. Le, and Jason Wei.
\newblock Scaling instruction-finetuned language models.
\newblock \emph{J. Mach. Learn. Res.}, 25:\penalty0 70:1--70:53, 2024.

\bibitem[Deng et~al.(2023)Deng, Gu, Zheng, Chen, Stevens, Wang, Sun, and Su]{deng2023mind2webgeneralistagentweb}
Xiang Deng, Yu~Gu, Boyuan Zheng, Shijie Chen, Samual Stevens, Boshi Wang, Huan Sun, and Yu~Su.
\newblock Mind2web: Towards a generalist agent for the web.
\newblock In \emph{NeurIPS}, 2023.

\bibitem[Gur et~al.(2024)Gur, Furuta, Huang, Safdari, Matsuo, Eck, and Faust]{gur2023real}
Izzeddin Gur, Hiroki Furuta, Austin~V. Huang, Mustafa Safdari, Yutaka Matsuo, Douglas Eck, and Aleksandra Faust.
\newblock A real-world webagent with planning, long context understanding, and program synthesis.
\newblock In \emph{{ICLR}}. OpenReview.net, 2024.

\bibitem[He et~al.(2024)He, Yao, Ma, Yu, Dai, Zhang, Lan, and Yu]{he2024webvoyager}
Hongliang He, Wenlin Yao, Kaixin Ma, Wenhao Yu, Yong Dai, Hongming Zhang, Zhenzhong Lan, and Dong Yu.
\newblock Webvoyager: Building an end-to-end web agent with large multimodal models.
\newblock In \emph{{ACL} {(1)}}, pp.\  6864--6890. Association for Computational Linguistics, 2024.

\bibitem[He et~al.(2021)He, Liu, Gao, and Chen]{he2021debertadecodingenhancedbertdisentangled}
Pengcheng He, Xiaodong Liu, Jianfeng Gao, and Weizhu Chen.
\newblock Deberta: decoding-enhanced bert with disentangled attention.
\newblock In \emph{{ICLR}}. OpenReview.net, 2021.

\bibitem[Kong et~al.(2024)Kong, Ruan, Chen, Zhang, Bao, Shi, Qing, Hu, Mao, Li, Zeng, Zhao, and Wang]{kong2023tptu}
Yilun Kong, Jingqing Ruan, Yihong Chen, Bin Zhang, Tianpeng Bao, Shiwei Shi, Du~Qing, Xiaoru Hu, Hangyu Mao, Ziyue Li, Xingyu Zeng, Rui Zhao, and Xueqian Wang.
\newblock Tptu-v2: Boosting task planning and tool usage of large language model-based agents in real-world industry systems.
\newblock In \emph{{EMNLP} (Industry Track)}, pp.\  371--385. Association for Computational Linguistics, 2024.

\bibitem[Li et~al.(2022)Li, Xu, Cui, and Wei]{li2021markuplm}
Junlong Li, Yiheng Xu, Lei Cui, and Furu Wei.
\newblock Markuplm: Pre-training of text and markup language for visually rich document understanding.
\newblock In \emph{{ACL} {(1)}}, pp.\  6078--6087. Association for Computational Linguistics, 2022.

\bibitem[Liu et~al.(2018)Liu, Guu, Pasupat, Shi, and Liang]{liu2018reinforcement}
Evan~Zheran Liu, Kelvin Guu, Panupong Pasupat, Tianlin Shi, and Percy Liang.
\newblock Reinforcement learning on web interfaces using workflow-guided exploration.
\newblock In \emph{{ICLR} (Poster)}. OpenReview.net, 2018.

\bibitem[Lu et~al.(2024)Lu, Kasner, and Reddy]{lù2024weblinxrealworldwebsitenavigation}
Xing~Han Lu, Zdenek Kasner, and Siva Reddy.
\newblock Weblinx: Real-world website navigation with multi-turn dialogue.
\newblock In \emph{{ICML}}. OpenReview.net, 2024.

\bibitem[Ma et~al.(2023)Ma, Zhang, Wang, Pan, and Yu]{ma2023laser}
Kaixin Ma, Hongming Zhang, Hongwei Wang, Xiaoman Pan, and Dong Yu.
\newblock {LASER:} {LLM} agent with state-space exploration for web navigation.
\newblock \emph{CoRR}, abs/2309.08172, 2023.

\bibitem[Mazumder \& Riva(2021)Mazumder and Riva]{mazumder2020flin}
Sahisnu Mazumder and Oriana Riva.
\newblock {FLIN:} {A} flexible natural language interface for web navigation.
\newblock In \emph{{NAACL-HLT}}, pp.\  2777--2788. Association for Computational Linguistics, 2021.

\bibitem[Pan et~al.(2024{\natexlab{a}})Pan, Zhang, Tomlin, Zhou, Levine, and Suhr]{pan2024autonomous}
Jiayi Pan, Yichi Zhang, Nicholas Tomlin, Yifei Zhou, Sergey Levine, and Alane Suhr.
\newblock Autonomous evaluation and refinement of digital agents.
\newblock \emph{CoRR}, abs/2404.06474, 2024{\natexlab{a}}.

\bibitem[Pan et~al.(2024{\natexlab{b}})Pan, Kong, Zhou, Cui, Leng, Jiang, Liu, Shang, Zhou, Wu, and Wu]{pan2024webcanvas}
Yichen Pan, Dehan Kong, Sida Zhou, Cheng Cui, Yifei Leng, Bing Jiang, Hangyu Liu, Yanyi Shang, Shuyan Zhou, Tongshuang Wu, and Zhengyang Wu.
\newblock Webcanvas: Benchmarking web agents in online environments.
\newblock \emph{CoRR}, abs/2406.12373, 2024{\natexlab{b}}.

\bibitem[Shaw et~al.(2023)Shaw, Joshi, Cohan, Berant, Pasupat, Hu, Khandelwal, Lee, and Toutanova]{shaw2023pixels}
Peter Shaw, Mandar Joshi, James Cohan, Jonathan Berant, Panupong Pasupat, Hexiang Hu, Urvashi Khandelwal, Kenton Lee, and Kristina~N Toutanova.
\newblock From pixels to ui actions: Learning to follow instructions via graphical user interfaces.
\newblock \emph{Advances in Neural Information Processing Systems}, 36:\penalty0 34354--34370, 2023.

\bibitem[Zheng et~al.(2023)Zheng, Wang, and An]{zheng2023synapse}
Longtao Zheng, Rundong Wang, and Bo~An.
\newblock Synapse: Leveraging few-shot exemplars for human-level computer control.
\newblock \emph{CoRR}, abs/2306.07863, 2023.

\end{thebibliography}
\bibliographystyle{colm2025_conference}

\appendix
\section{Common Prior Workflow}
\label{sec:appendix}
\begin{figure*}[]
    \centering \includegraphics[width=\textwidth]{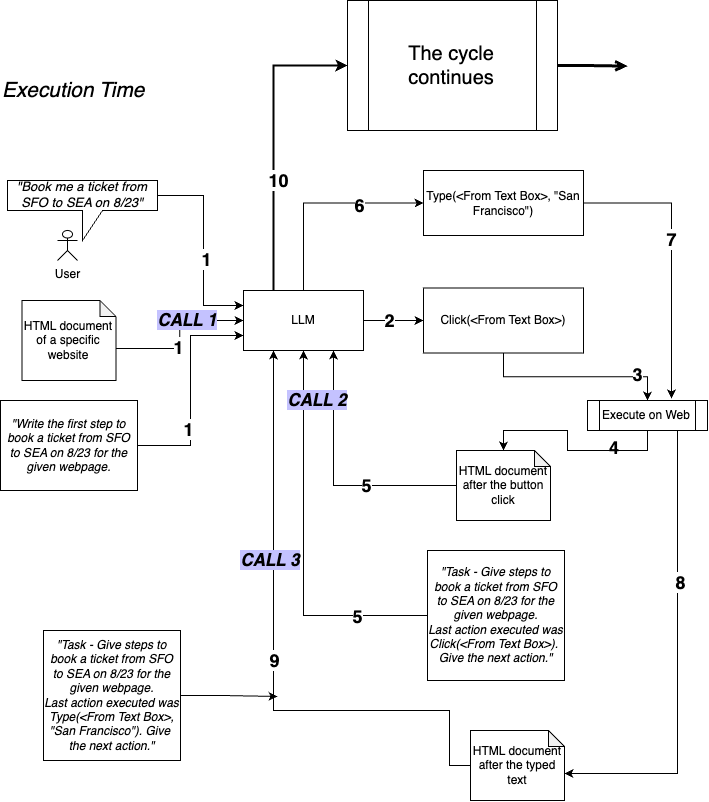}
    \caption{Common workflow of existing solutions like MindAct~\citep{deng2023mind2webgeneralistagentweb}.}
    \label{fig:other_options}
\end{figure*}

Existing solutions iteratively call the LLM after every single action, which is inefficient (see Figure~\ref{fig:other_options}), and does not take into account planning that Agents can leverage to understand the web like a human would, by constructing viable workflows and following those action sequences. Previously executed tasks \textit{should} inform similar future tasks. 

\section{Action Library Workflow}
\label{sec:applib}
\begin{figure*}[htbp]
    \centering \includegraphics[width=\columnwidth]{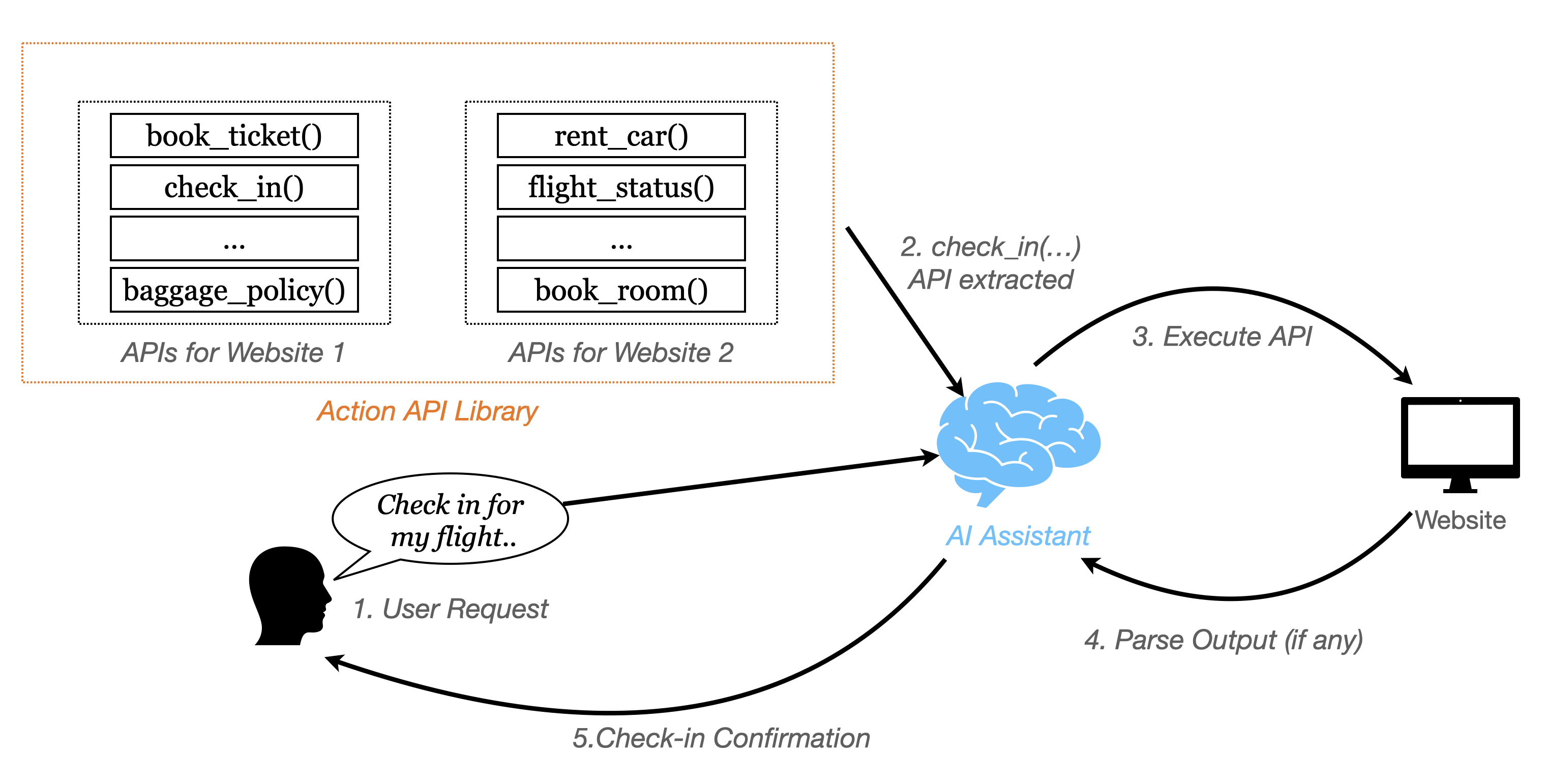}
    \caption{Using Action Library.}
    \label{fig:step4}
\end{figure*}

The Figure~\ref{fig:step4} depicts PAFFA's core inference-time process. Instead of parsing HTML, the system retrieves a relevant pre-computed API from the library based on the user's request, extracts parameters, and executes the API directly to interact with the website.
\section{Dataset}
\label{subsec:dataset_app}
We evaluate our algorithm on the Mind2Web benchmark~\citep{deng2023mind2webgeneralistagentweb}, which provides annotated actions for diverse tasks across real-world websites. This dataset's key advantage is its use of complex, real-world websites rather than simplified environments like Mini-WoB++~\citep{liu2018reinforcement}, enabling more realistic evaluation of generalization capabilities.

For our evaluation, we extract a focused subset of Travel-Airlines domain and Shopping-General Cross-Task splits. We additionally include the Cross-Website split of Shopping-General to assess generalization in our no-train setting (see Appendix \ref{subsec:appendix_data} for more details). From each website's training data, we extract 'unique web pages' - pages with distinct HTML structures whose content may change dynamically (e.g., a homepage's structure remains constant while its content varies with date and location).

We note two key limitations of the benchmark: lack of interactivity compared to toy environments, and its evaluation methodology requiring exact matching of user actions. As the dataset expects models to map HTML documents, previous actions, and tasks to single-step actions, it presents a potential disadvantage for our approach, which aims to avoid task-specific implementations and live HTML parsing at test time.

\subsection{Dataset Details}
\label{subsec:appendix_data}
The three datasets are:
\begin{enumerate}
    \item Cross-Task Split for Travel-Airlines : contains 31 test tasks, across 7 websites : American Airlines, Delta, Jetblue, Kayak, Qatarairways, Ryanair, United.
    \item Cross-Task Split for Shopping-General : contains 8 test tasks, across 3 websites : Amazon, Target, Instacart.
    \item Cross-Website Split for Shopping-General : contains 17 test tasks, for one website : Google Shopping.
\end{enumerate}

Note : There are no Cross-Website test cases for Travel-Airlines.

\section{Qualitative Evaluation Metrics}
\label{sec:qualapp}
We employed LLM-based evaluation (using Sonnet 3.5) to assess the quality of initially generated scripts along the following dimensions:
\begin{itemize}
    \item Script Task Alignment (Scale 1-5): Measures how completely the generated script fulfills all explicit and implicit requirements of the specified task, comparing against the user instruction and available expert action sequences used solely as an evaluation reference. Crucially, similar to the `Inexact' evaluation in Section~\ref{subsec:bench}, this metric prioritizes functional goal completion. A high score indicates the script successfully achieves the task's objectives, even if the sequence of actions differs from a specific reference path. A lower score indicates failure to meet requirements or incomplete execution.
    \item Action Representation Fidelity (Scale 1-5): Assesses how accurately the essential actions and parameters within the generated script capture the necessary steps required by the task, reflecting the faithfulness of the generated code structure to the underlying interaction logic.
    \item Script Efficiency (Scale 1-5): Assesses whether the script uses a reasonably direct path, avoiding unnecessary steps or redundant operations. A higher score indicates greater efficiency.
\end{itemize}

\section{Example Action Generation and Usage}
\label{app:api_example}
This appendix provides a concrete example of the Action API Library components for the Delta website, illustrating the output of the API Synthesis process and how APIs are used.

\subsection{Example Synthesized APIs for Delta}
Following the process described in Section~\ref{subsec:lib}, scripts generated for various Delta tasks (e.g., finding reservations, searching for flights using different criteria) are clustered and synthesized into parameterized Python functions. Below are two examples of such APIs generated for the Delta website:

\begin{itemize}
    \item API for Retrieving Trip Information: This function encapsulates the steps needed to look up a reservation using confirmation details.
    \begin{lstlisting}[language=Python, label={lst:trip_info_delta}]

def retrieve_trip_information(
    driver: webdriver.Chrome,
    confirmation_number: str,
    first_name: str,
    last_name: str,
    wait_time: int = 10
) -> None:
    """
    Retrieves trip information by navigating to the 'My Trips' page, 
    entering the confirmation number, first name, and last name, 
    then submitting the form.
    
    Args:
        driver (webdriver.Chrome): The Selenium WebDriver instance.
        confirmation_number (str): The trip confirmation number.
        first_name (str): The traveler's first name.
        last_name (str): The traveler's last name.
        wait_time (int, optional): Max wait time. Defaults to 10.
    """
    def safe_action(locator: tuple, action: str, value: Optional[str] 
    = None) -> None:
        element = WebDriverWait(driver, wait_time).until(
            EC.presence_of_element_located(locator)
        )
        if action == "click":
            driver.execute_script("arguments[0].click();", element)
        elif action == "input":
            # Use arguments array for safer JS execution with values
            driver.execute_script("arguments[0].value=arguments[1];"
            , element, value)

    # Navigate to My Trips
    safe_action((By.ID, "headPrimary3"), "click")
    sleep(2) # Consider replacing sleep with WebDriverWait if possible

    # Enter details
    safe_action((By.ID, "confirmationNo"), "input", confirmation_number)
    safe_action((By.ID, "firstName"), "input", first_name)
    safe_action((By.ID, "lastName"), "input", last_name)

    # Submit the form
    safe_action((By.ID, "btn-mytrip-submit"), "click")
    sleep(3) # Wait for potential results page
    \end{lstlisting}
    \item API for Searching Flights: This function handles various flight search scenarios, including booking, checking status, using miles, and specifying class.
    \begin{lstlisting}[language=Python,label={lst:book_delta}]
    def search_flights(
    driver: webdriver.Chrome,
    origin: str,
    destination: str,
    depart_date: str, # Expects format like MM/DD/YYYY
    use_miles: bool = False,
    wait_time: int = 10
) -> None:
    """
    Searches for flights using origin, destination, date, and miles option.

    Args:
        driver (webdriver.Chrome): Selenium WebDriver instance.
        origin (str): Origin airport/city.
        destination (str): Destination airport/city.
        depart_date (str): Departure date (e.g., "06/05/2024").
        use_miles (bool, optional): Search using miles. Defaults to False.
        wait_time (int, optional): Max wait time. Defaults to 10.
    """
    def safe_action(locator: tuple, action: str, value: Optional[str] = None) -> None:
        element = WebDriverWait(driver, wait_time).until(
            EC.element_to_be_clickable(locator) 
            # Use clickable for inputs/buttons
        )
        if action == "click":
             # Using execute_script often more reliable than element.click()
            driver.execute_script("arguments[0].click();", element)
        elif action == "input":
            element.click() # Focus element first
            element.clear()
            element.send_keys(value)
            # May need Keys.RETURN or specific waits for autocomplete here
        sleep(0.5) # Small pause

    #  Enter Origin
    safe_action((By.ID, "fromAirportName"), "input", origin)
    sleep(1)

    #  Enter Destination
    safe_action((By.ID, "toAirportName"), "input", destination)
    sleep(1)

    #  Enter Departure Date
    safe_action((By.ID, "input_departureDate_1"), "input", depart_date)
    sleep(1)

    # Handle Shop with Miles
    if use_miles:
         safe_action((By.ID, "shopWithMiles"), "click")
         sleep(1)

    # Click Search Button
    safe_action((By.ID, "btn-book-submit"), "click")

    sleep(3) # Wait for search results page

    \end{lstlisting}
\end{itemize}

\subsection{Synthetic Task Generation and API Grounding}
To test the coverage and robustness of the generated API library beyond the initial tasks used for its creation, we also employed LLM-based synthetic task generation (zero-shot prompting with Sonnet 3.5). For a given website (e.g., Delta), we prompted the LLM to generate diverse, realistic user requests.
\textbf{Example Synthetically Generated Tasks for Delta}:
\begin{itemize}
    \item Find my reservation with confirmation code DLTX7Y including passenger name Sarah Johnson
    \item Check available non-stop flights from Atlanta to Los Angeles on September 12th 2025 for two adults in Comfort+
    \item Show me first class round-trip options from Boston to Miami departing April 15, 2026 and returning April 22, 2026
    \item Find flights from Chicago to San Francisco on July 8th 2025 that can be booked using SkyMiles and have Wi-Fi onboard
\end{itemize}

These synthetic tasks are then processed by PAFFA's runtime system (Section~\ref{subsec:runtime}). The LLM performs API Retrieval/Intent Recognition and Parameter Extraction/Slot Filling to map the natural language task to the most appropriate API call.

\textbf{Example Grounded API Call}: For a request like ``Find flights from Seattle to New York on June 5th, 2025 using miles'', the system might generate the following call:

\begin{lstlisting}[language=Python,label={lst:call}]
    
    search_flights(
        driver=driver_instance, 
        origin="Seattle", 
        destination="NewYork", 
        date="2025-06-05", 
        use_miles=True 
    # travel_class might be None if not specified
    )
\end{lstlisting}
This process of generating synthetic tasks and grounding them to API calls allows for broader testing of the library's capabilities and the LLM's ability to correctly map diverse requests to the pre-computed functions.

\end{document}